\documentclass[conference]{IEEEtran}
\IEEEoverridecommandlockouts
\usepackage{cite}
\usepackage{enumitem} % in the preamble
\usepackage[ruled,vlined]{algorithm2e}
\usepackage{mdframed}
\usepackage{amsmath,amssymb,amsfonts}
\usepackage{algorithmic}
\usepackage{graphicx}
\usepackage{hyperref}
\usepackage{float}
\usepackage{textcomp}
\usepackage{xcolor}
\usepackage{tikz}
\usepackage{multirow}  % Add this to preamble if missing
\usepackage{graphicx}  % For \resizebox
\usepackage{booktabs}  % Optional: better lines

\usepackage{cellspace} % add in your preamble
\def\BibTeX{{\rm B\kern-.05em{\sc i\kern-.025em b}\kern-.08em
    T\kern-.1667em\lower.7ex\hbox{E}\kern-.125emX}}

\usepackage[none]{hyphenat}
\begin{document}

\title{SA-ADP: Sensitivity-Aware Adaptive Differential Privacy for Large Language Models \\[5pt] 
\normalsize{\textbf{\textcolor{red}{Author's draft for soliciting feedback - \today}}}}

\author{
    \IEEEauthorblockN{Stella Etuk}
    \IEEEauthorblockA{\textit{School of Information Technology} \\
    \textit{Carleton University} \\
    \textit{stellaetuk@cmail.carleton.ca}}
    \and
    \IEEEauthorblockN{Ashraf Matrawy}
    \IEEEauthorblockA{\textit{School of Information Technology} \\
    \textit{Carleton University} \\
    \textit{ashraf.matrawy@carleton.ca}}
}
\maketitle
\begin{abstract}

Despite advances in the use of large language models (LLMs) in downstream tasks, their ability to memorize information has raised privacy concerns. Therefore, protecting personally identifiable information (PII) during LLM training remains a fundamental challenge. Conventional methods like Differential Privacy-Stochastic Gradient Descent (DP-SGD) provide robust privacy protection via uniform noising, protecting PII regardless of its distinct sensitivity. This comes at the expense of the model's utility, leading to a trade-off. In this paper, we propose SA-ADP, a sensitivity-aware approach that allocates noise based on the sensitivity of individual PII. We evaluated our method on four datasets (ABCD, CUSTOMERSIM, Wikitext-2, and UNSW-NB15 ). Our results show that SA-ADP achieves results comparable to the baseline (No-DP) and the conventional DP-SGD. This means that our method did not degrade the model's utility while still maintaining strong privacy protection.
\end{abstract}

\section{Introduction}
Large Language Models (LLMs) such as GPT by OpenAI and others have significantly advanced the field of natural language processing \cite{36.wibawa2024advancements}, powering applications such as chatbots, content generation, translation, and more. These models are trained on a large corpus of real-world data \cite{37.ju2024training} that often contains Personally Identifiable Information (PII), such as names, email addresses, phone numbers, and other sensitive information. This poses a serious privacy risk, as LLMs can memorize and expose such private information during model inference \cite{38.singh2024whispered}.

The risk of PII leakage highlights the critical importance of privacy protection during LLM training and fine-tuning \cite{39.charles2024fine}. 
Differential Privacy (DP) offers a promising solution, with Differential Privacy Stochastic Gradient Descent (DP-SGD) \cite{4.abadi2016deep} being a widely used variant. DP-SGD adds calibrated noise to clipped gradients during training, thereby obscuring the contribution of any single data point and mitigating privacy risks. Although this technique improves privacy, it also introduces \textbf{a trade-off between the model's utility and privacy}, where excessive noise reduces model performance, whereas insufficient noise compromises privacy \cite{40.das2025revisiting}. 
\textbf{A further limitation} of standard DP methods is the uniform application of noise across all tokens, without considering differences in sensitivity. 
For instance, a Social Security Number (SSN) poses far greater risk than a first name, yet both are treated equally under conventional DP-SGD. 
This approach results in inefficiencies in the privacy--utility trade-off. 

To address this limitation, we propose a \ textbf { Sensitivity-Aware Adaptive Differential Privacy (SA-ADP)} mechanism that adaptively adds noise to gradients based on token sensitivity. This proposed approach follows a three-stage pipeline: Detection, Scoring, and Noising. First, we use PII detection algorithms to detect sensitive tokens within the training data. Next, each token is assigned a normalized sensitivity score (0.0--1.0) based on its frequency, linkability, and datatype classification. Finally, we apply per-token adaptive noise during gradient clipping, with high-sensitivity tokens receiving stronger noise (0.51 - 1.00) and low noise to tokens classified as low sensitivity (0.01 - 0.50). The Rényi Differential Privacy (RDP) accountant is then used to track and bound the cumulative privacy loss budget throughout the training process, thereby guaranteeing privacy \cite{49.mironov2017renyi}.

This paper makes the following \textbf{contributions}:

\begin{enumerate}[leftmargin=*, labelsep=0.5em]
    \item[1a)] We propose Sensitivity-Aware Adaptive Differential Privacy (SA-ADP), a novel framework for LLM fine-tuning that adjusts noise levels according to token-level PII sensitivity.
    \item[1b)] We design an adaptive Detection–Scoring–Noising pipeline that introduces token-level sensitivity scoring based on frequency, linkability, and data type classification, enabling fine-grained, adaptive privacy protection.
    \item[2)] We implement and evaluate SA-ADP on four datasets (UNSW, ABCD, Wikitext-2, and CUSTOMERSIM dataset), demonstrating that our token-level adaptive approach maintains a strong privacy level while preserving model utility across diverse data scenarios.
\end{enumerate}

\section{Related Work}

\textbf{\textit{Privacy Risks in LLM}}:  Recent studies have shown that LLMs can memorize and reproduce sensitive content from their training data \cite{23carlini2019secret}, particularly when trained on large-scale, unfiltered datasets. This has also raised serious concerns about the development and deployment of LLMs in sensitive sectors like healthcare and financial institutions. LLMs can infer personal attributes from user input beyond memorized data \cite{1.staab2023beyond}. To mitigate these risks, researchers have proposed various methods, such as Differential Privacy (DP) \cite{25dwork2006calibrating}, data anonymisation for text data \cite{56lison2021anonymisation}, and data scrubbing with machine unlearning \cite{47.ren2025sok}, self-normalizing DP methods as proposed by Ibitoye et al.  \cite{10.ibitoye2022differentially}, DP with stochastic gradient clipping  (DP-SGD)  \cite{4.abadi2016deep}, all in an attempt to preserve the privacy of PIIs. While DP-SGD operates at the per-sample or per-record level, clipping each sample’s gradient and applying uniform Gaussian noise to the aggregated gradient, our approach focuses on per-token adaptive noise injection based on PII sensitivity, enabling fine-grained privacy protection. 

\vspace{0.5em}

\textbf{\textit{Differential Privacy}}: Differential Privacy (DP) introduced by Dwork \cite{41.dwork2006differential} is a framework that ensures that the presence or absence of any individual data point does not affect the output. Building on this, Abadi et al. \cite{4.abadi2016deep} proposed "Differential Privacy Stochastic Gradient Descent" (DP-SGD), which applies per-sample gradients clipping and adds Gaussian noise uniformly to the gradients during model training. This framework has become one of the most prominent techniques for data preservation. While these methods ensured privacy, the trade-off was the model's utility. Since then, different variants of the DP-SGD method have surfaced to train efficiently \cite{53.shi2022just} or improve noise calibration \cite{38.singh2024whispered}, \cite{59shi2021selective} in an attempt to enhance model utility while maintaining privacy. Still, none of these methods have considered the unique sensitivity of each PIIs token contained in the training data before adding noise.

\section{The Proposed Sensitivity-Aware Adaptive Differential Privacy Model (SA-ADP)}
This section presents the architectural and algorithmic components of our proposed SA-ADP framework for privacy-preserving fine-tuning of LLMs, with a primary focus on balancing privacy and utility. 

\begin{figure}[t]
\centering
\includegraphics[width=1.05\linewidth]{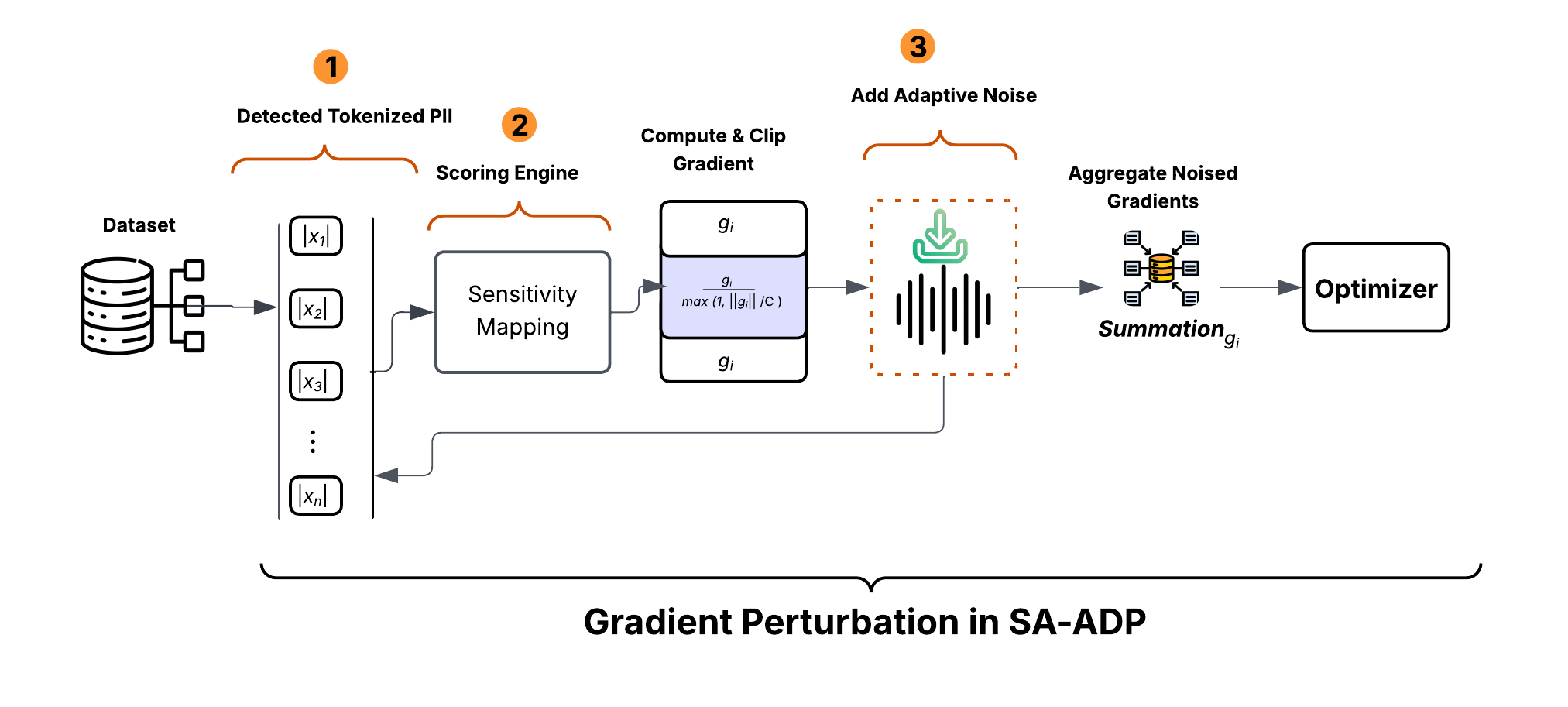}
\caption{Gradient Perturbation in the SA-ADP Framework. At \textbf{\textcolor{orange}{1}}, PII are detected from the input data and passed to the scoring engine at \textbf{\textcolor{orange}{2}}, where each token is assigned a sensitivity score. An adaptively calibrated noise multiplier is then computed and used at \textbf{\textcolor{orange}{3}} to inject Gaussian noise adaptively into the clipped gradients \cite{4.abadi2016deep} during the backward pass, ensuring stronger perturbation for high-sensitivity tokens and minimal distortion for low-sensitivity ones.}

\label{fig:gradient_perturbation}
\end{figure}

As shown in Fig. \ref{fig:gradient_perturbation}, we propose \textit{Sensitivity Aware Adaptive Differential Privacy (SA-ADP)}, which introduces token-specific privacy modulation based on PII sensitivity. The mechanism is built by introducing three core steps into the training pipeline: PII detection, sensitivity scoring, and adaptive noise allocation. The following subsections explain each core of our proposed method.

\subsection{PII Detection}

We utilize a LangChain-based privacy agent powered by LLaMA 3 (via Ollama) \cite{LangChain2023,Grattafiori2024Llama3,Lin2025JOSS}to
automatically extract PII elements from structured text records. Each record is passed to the
agent with a task-specific prompt, and the model returns a structured array listing PII
fields and values. This detection process is applied over a corpus of dataset entries.

\subsection{ Sensitivity Scoring}
\label{sec:sensitivity_scoring}
\begin{figure}[!htbp]
\centering
\includegraphics[width=1.0\linewidth]{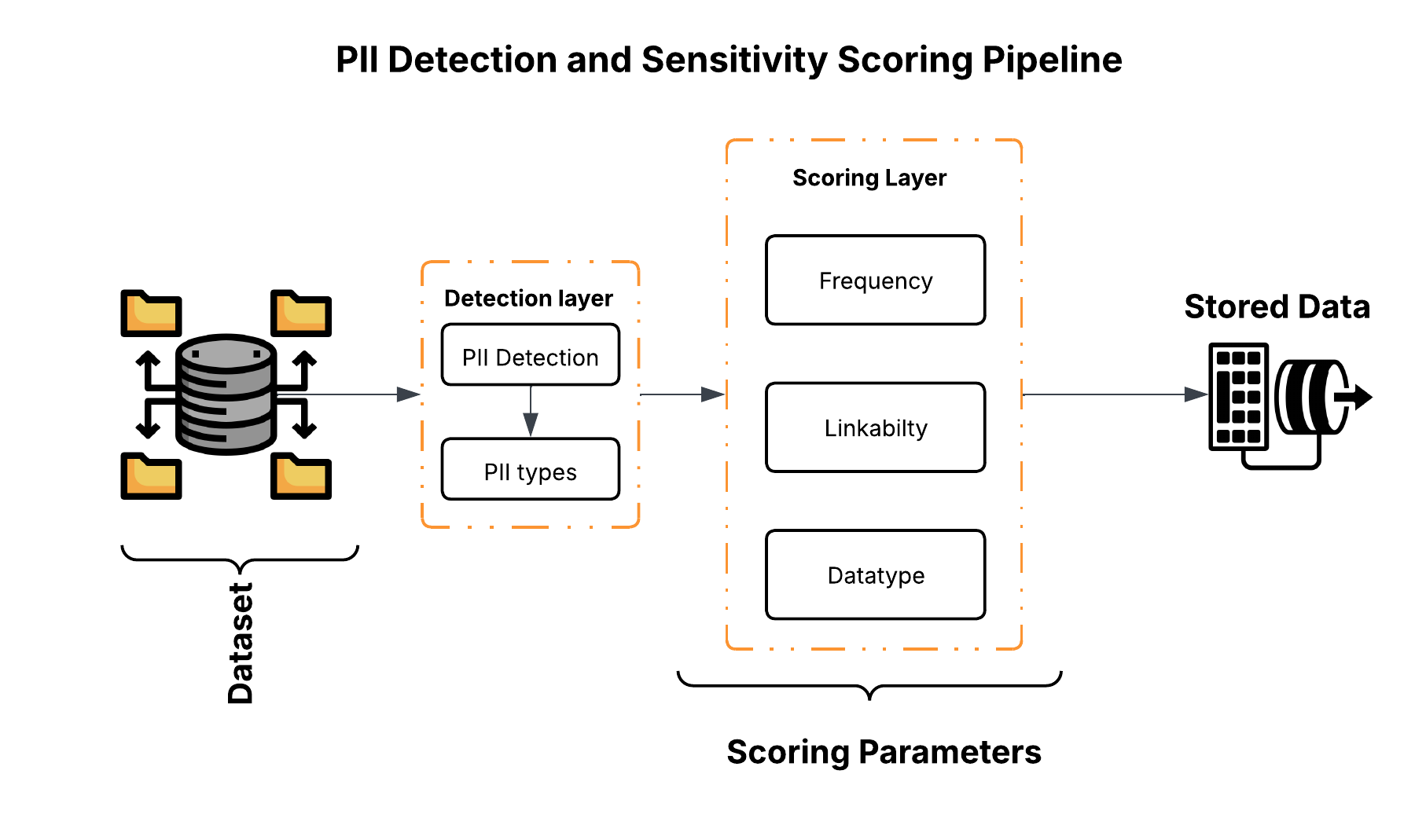}
\caption{PII Detection and Sensitivity Scoring. A detection agent processes raw input to identify PII tokens. Each identified token is scored based on frequency, linkability, and datatype parameters. The resulting sensitivity score is used to inform adaptive privacy mechanisms.}
\label{fig:pii_pipeline}
\end{figure}

To enable token-level adaptive differential privacy, we first identify and determine the sensitivity levels of the different PII types (\( p_i \) ) in the training data. This involves detecting PII types and assigning each a sensitivity score based on three dimensions (Frequency, Linkability, and Datatype) that could reflect the risk level of a PII type \cite{NIST200-122mccallister2010guide}. Fig. \ref{fig:pii_pipeline} illustrates the process used for this analysis.

\subsubsection{Scoring Dimension}

Each detected PII token is scored along three dimensions:

\begin{itemize}
    \item \textbf{Frequency Score (\( S_{\text{freq}} \))}: This measures the rarity of PII type in the dataset. 
    Let \( p_i \) denote the $i$-th detected PII  type in the dataset and let $P = \{p_1, p_2, \dots, p_n\}$ represent the complete set of detected PII types.
    \begin{equation}
        S_{\text{freq}}(p_i) = 1 - \frac{f(p_i)}{N}
    \end{equation}
    where \( f(p_i) \) is the count of each PII type \( p_i \), and \( N \) is the total number  PII count in the dataset.

    \item \textbf{Linkability Score (\( S_{\text{link}} \))}: Indicates how likely a PII type can be linked with others. This is assessed by querying an LLM agent:
    \begin{equation}
        S_{\text{link}}(p_i) = 
        \begin{cases}
            1, & \text{if PII type is linkable} \\
            0, & \text{otherwise}
        \end{cases}
    \end{equation}

    \item \textbf{DataType Classification Score (\( S_{\text{datatype}} \))}: Determines whether the PII type is legally recognized under global privacy laws (e.g., GDPR, HIPAA):
    \begin{equation}
        S_{\text{datatype}}(p_i) =
        \begin{cases}
            1, & \text{if PII type is datatype protected} \\
            0, & \text{otherwise}
        \end{cases}
    \end{equation}
\end{itemize}

\subsubsection{Aggregated Privacy Sensitivity Index}
%\label{sec:sensitivity_scoring}
The three scores are combined into a unified privacy sensitivity index \( S_{\text{final}} \in [0, 1] \) through a weighted linear combination:

\[
S_{\text{final}}(p_i) = W_1 \cdot S_{\text{freq}}(p_i) + W_2 \cdot S_{\text{link}}(p_i) + W_3 \cdot S_{\text{datatype}}(p_i)
\tag{4}
\]

The weights were selected to emphasize rarity while balancing contextual and regulatory importance. In this case, we set: frequency = 0.4, linkability = 0.3, and datatype = 0.3

\subsubsection{Algorithm 1: Algorithmic Implementation}

\begin{algorithm}
\caption{PII Sensitivity Scoring}
\label{alg:sensitivity_scoring}
\KwIn{Dataset \( D \), Detected PII set \( P \)}
\KwOut{Sensitivity score mapping \( S_{\text{final}}(p_i) \) for all \( p_i \in P \)}

Compute total PII count: \( N \gets \text{count\_words}(D) \)\;

\ForEach{PII type \( p \in P \)}{
    Compute frequency score: \( S_{\text{freq}}(p) \gets 1 - \frac{f(p)}{N} \)\;
    Query LLM for linkability score: \( S_{\text{link}}(p) \in \{0, 1\} \)\;
    Query LLM for datatype classification score: \( S_{\text{datatype}}(p) \in \{0, 1\} \)\;
    Compute final sensitivity score:\;
    \qquad \( S_{\text{final}}(p) \gets 0.4 \cdot S_{\text{freq}}(p) + 0.3 \cdot S_{\text{link}}(p) + 0.3 \cdot S_{\text{datatype}}(p) \)\;
}
\end{algorithm}

\subsection{Noise Calibration Strategy}

Since the privacy guarantee in Differential Privacy (DP) is defined by the budget parameter \(\epsilon\),  DP algorithms regulate privacy by introducing a unitless Gaussian noise \( \sigma \), which is calibrated inversely to \(\epsilon\) into the training data. \cite{64.nanayakkara2023chances}. To enable fine-grained privacy control, our approach adaptively calibrates \( \sigma \)  on a per-token basis ( for each PII type) using the computed sensitivity scores \( S_{\text{final}(p_i)} \in [0,1] \) in equation 4. We have that:

\[
\epsilon = \frac{C}{\sigma} \tag{5}
\]

(where C is the clipping norm as proposed in the framework of  Abadi et al. \cite{4.abadi2016deep}

The noise calibration strategy  introduces a deterministic mapping function \( \mathcal{F} \colon S_{\text{final}(p_i)} \rightarrow \sigma \), which assigns each token a Gaussian noise scale according to its sensitivity:

\[
\sigma = 
\begin{cases}
\sigma_{\text{low}}, & \text{if } 0.01 \leq S_{\text{final}(p_i)} \leq 0.50 \\
\sigma_{\text{high}}, & \text{if } 0.51 \leq S_{\text{final}(p_i)} \leq 1.00 \\
0, & \text{otherwise}
\end{cases}
\tag{6}
\]

Where \( \sigma_{\text{low}} \) and \( \sigma_{\text{high}} \) are adjustable noise calibration constants (in our experiments, 2.0 and 3.0, respectively). This mapping ensures that higher-sensitivity tokens are injected with stronger noise, while non-sensitive tokens are perturbed minimally, thus preserving model utility.

\vspace{0.2em}

\begin{algorithm}
\SetAlgoLined
\caption{Token-wise Noise Injection}
\label{alg:noise_injection}
\KwIn{Token sequence \( T = [t_1, \dots, t_n] \), Sensitivity scores \( S = [s_1, \dots, s_n] \), gradient vector \(G = [g_1, \dots, g_n]\), noise map \(\mathcal{M}\)}
\KwOut{Noised gradient vector \(\hat{G}\)}

\For{\(i \leftarrow 1\) \KwTo \(n\)}{
    \(\sigma_i \leftarrow \mathcal{M}(s_i)\) \tcp*{Map score to noise level}
    \(\hat{g}_i \leftarrow \text{clip}(g_i, C)\) \tcp*{Clip gradient norm}
    \If{\(\sigma_i > 0\)}{
        \(\hat{g}_i \leftarrow \hat{g}_i + \mathcal{N}(0, \sigma_i^2 C^2 I)\) \tcp*{Add Gaussian noise Adapted from the core DP-SGD mechanism by Abadi et al. \cite{4.abadi2016deep}}
    }
    Store \(\hat{g}_i\) in \(\hat{G}\)
}
\textbf{return} \(\hat{G}\) \\

\end{algorithm}

\section{Experiments}
\textit{\textbf{Datasets}}: To evaluate the effectiveness of SA-ADP, we fine-tune the GPT-2 base model on both unstructured and structured datasets. For the unstructured dataset, we used the \textbf{Wikitext-2 Dataset}, a natural language corpus commonly used as a benchmark for language model performance \cite{67wiki.merity2016pointer}. We also considered two synthetic conversational datasets, which include the \textbf{ABCD Dataset} introduced by Shi et al. \cite{53.shi2022just} and the \textbf{CustomerSim Dataset} used in \cite{59shi2021selective}. Finally, we used the \textbf{UNSW-NB15} network traffic dataset \cite{65.moustafa2015unsw} for structured data. This coverage ensures that the framework is empirically validated across diverse data modalities where privacy sensitivity and linguistic utility interact in different ways.

\textbf{\textit{Model Fine-tuning with SA-ADP}}: We fine-tune a pre-trained \textbf{GPT-2} model using our SA-ADP framework. The training uses the Adam optimizer with a constant learning rate \( \eta = 0.001 \), a batch size of 16, and a sequence length of 512 over three epochs with a sample rate $q = 0.1$. Differential privacy is achieved by applying adaptive Gaussian noise \( \sigma_i \in [3.0, 2.0] \) and gradient clipping (\( C = 1.0 \)) with stronger noise assigned to more sensitive tokens and minimal perturbation for less sensitive ones. We also implemented the DP-SGD framework under the same hyperparameter tuning for comparison.

\textbf{\textit{Privacy Accounting}}: To quantify the privacy guarantees of the method, we use Rényi Differential Privacy (RDP) as our accounting framework, following the formulation introduced by Mironov \cite{49.mironov2017renyi}. This approach enables cumulative tracking of privacy loss budget (\( \epsilon \)) across training steps. We use a fixed Rényi order \( \alpha = 32 \) and a target delta \( \delta = 10^{-5} \) for all runs. The final privacy levels are computed using the standard RDP conversion \cite{49.mironov2017renyi}

\[
    \epsilon = \min_{\alpha > 1} \left\{ \epsilon_{\text{RDP, total}}(\alpha) 
    + \frac{\log(1/\delta)}{\alpha - 1} \right\}.
    \tag{7}
\]

\section{Results and Evaluation}

This section presents the performance of our proposed SA-ADP framework and the DP-SGD framework, evaluated and compared using accuracy, perplexity, and privacy budgeting metrics.

\subsection{\textbf{Evaluation Metrics}}

We evaluate the effectiveness of the proposed SA-ADP mechanism using two key dimensions.

\textit{Utility}: To comprehensively evaluate the proposed SA-ADP mechanism, we use both utility- and privacy-oriented metrics. Utility is measured through classification accuracy and language modeling perplexity. Accuracy reflects the model’s ability to correctly classify or predict task-specific labels, particularly relevant for structured datasets such as UNSW-NB15. Perplexity measures how well the model predicts token sequences, indicating fluency and generalization; lower values signal outputs closer to natural human language, which is essential for conversational datasets such as ABCD.

\textit{Privacy}: Privacy is quantified through the privacy loss budget parameter $(\epsilon, \delta)$ under the Rényi Differential Privacy (RDP) framework. A lower value of indicates stronger privacy levels, meaning the model's output is less influenced by any single training example. We adopted a fixed \( \delta = 10^{-5} \) for all experiments to allow consistent comparisons between models.

The detailed evaluation metrics of our SA-ADP and DP-SGD mechanisms across datasets are summarized in Table~\ref{tab:appendix_results}.

\begin{table*}[!htbp]
\centering
\caption{Comparison of Baseline (No DP), SA-ADP, and DPSGD Across Datasets}
\label{tab:appendix_results}
\renewcommand{\arraystretch}{1.5}
\resizebox{\textwidth}{!}{%
\begin{tabular}{|l|c|c|c|c|c|c|c|c|c|c|c|}
\hline
\multirow{3}{*}{\textbf{Datasets}} 
& \multicolumn{2}{c|}{\textbf{Baseline (No DP)}} 
& \multicolumn{3}{c|}{\textbf{SA-ADP}} 
& \multicolumn{3}{c|}{\textbf{DPSGD $\sigma$=2}} 
& \multicolumn{3}{c|}{\textbf{DPSGD $\sigma$=3}} \\ 
\cline{2-12}
& \textbf{Accuracy} & \textbf{Perplexity} 
& \textbf{Accuracy} & \textbf{Perplexity} & \textbf{Epsilon} 
& \textbf{Accuracy} & \textbf{Perplexity} & \textbf{Epsilon} 
& \textbf{Accuracy} & \textbf{Perplexity} & \textbf{Epsilon} \\ 
\hline
\textbf{UNSW-NB15} & 76.67 & 1.05 & 76.92 & 1.05 & 0.1869 & 76.48 & 3.33 & 1.7246 & 5.82 & 8.22 & 1.0712 \\ 
\hline
\textbf{ABCD} & 97.69 & 1.03 & 97.68 & 1.03 & 0.1858 & 97.73 & 1.09 & 1.5039 & 97.72 & 1.10 & 0.9401 \\ 
\hline
\textbf{Wikitext-2} & 62.55 & 2.95 & 60.96 & 3.05 & 0.7697 & 38.12 & 20.17 & 2.7711 & 25.01 & 133.51 & 1.1259 \\ 
\hline
\textbf{CUSTOMERSIM} & 94.20 & 1.18 & 94.22 & 1.18 & 0.6399 & 94.20 & 1.38 & 2.5096 & 94.21 & 1.51 & 1.5319 \\ 
\hline
\end{tabular}%
}
\end{table*}

\subsection{\textbf{Results and Discussion}}

\begin{figure}[!htbp]
    \centering
    \includegraphics[width=0.90\linewidth]{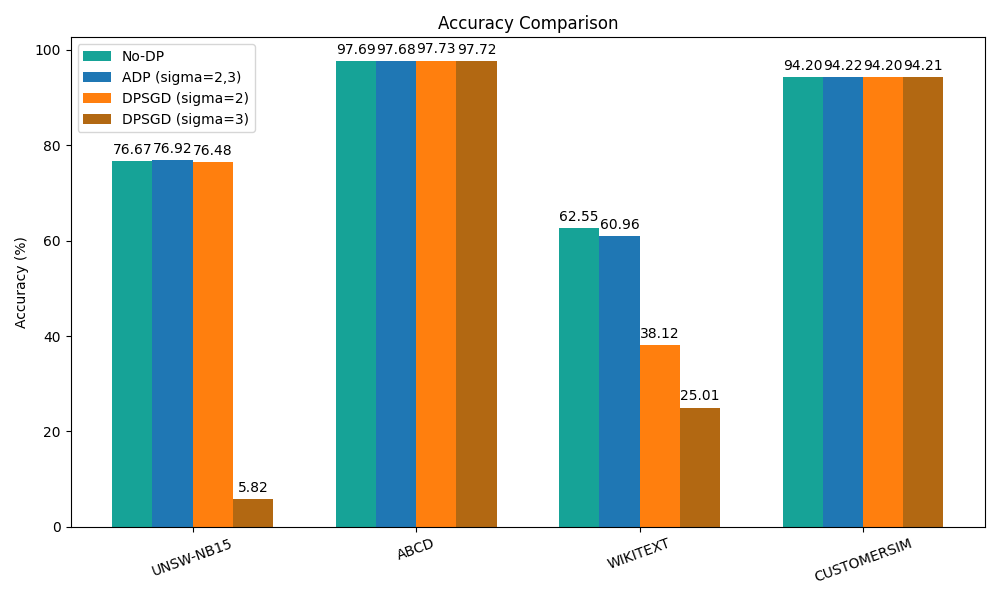}
    \caption{Accuracy Comparison between SA-ADP and DP-SGD across all three Datasets.}
    \label{fig:accuracy_comparison}
\end{figure}

\begin{figure}[!htbp]
    \centering
    \includegraphics[width=0.90\linewidth]{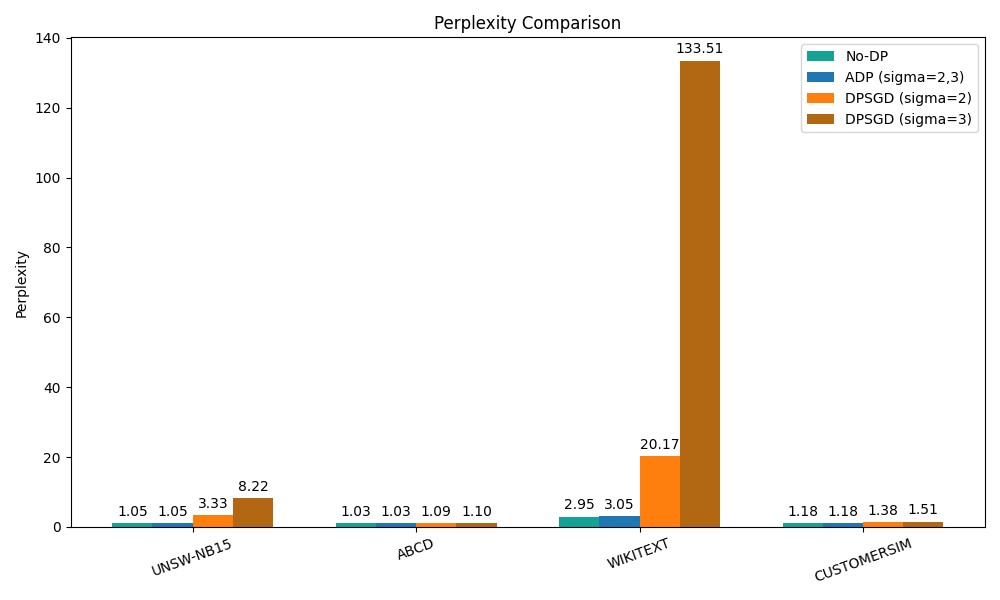}
    \caption{Perplexity Comparison between SA-ADP and DP-SGD across all three Datasets. Lower perplexity indicates better fluency.}
    \label{fig:perplexity_comparison}
\end{figure}

\textit{\textbf{Privacy-Utility Trade-offs with SA-ADP:}}
We evaluate the SA-ADP framework across multiple benchmark datasets with varying PII density (the proportion of tokens in a dataset that are identified as PII, in contrast to datasets where most tokens are non-PII). Compared with the baseline (No DP), SA-ADP maintains comparable performance in both accuracy and perplexity (Fig. \ref{fig:accuracy_comparison} for the accuracy comparison and Fig. \ref{fig:perplexity_comparison} for the perplexity comparison), indicating that the SA-ADP framework does not degrade the model. Compared to DP-SGD, SA-ADP slightly outperforms it in perplexity (lower is better) while maintaining a low privacy loss budget ($\epsilon$) across all datasets (See Table~\ref{tab:appendix_results}). This is particularly evident in conversational datasets (ABCD and CustomerSim) with low PII density. For the Wikitext-2 benchmark, which has a higher density of PII, a similar trend is observed, as the model does not suffer significant degradation compared to the baseline despite an increase in its $\epsilon$ value. 

\textbf{\textit{Consistency of SA-ADP Across Structured and Unstructured Data: }}
Our framework was tested across datasets spanning several domains (Structured and unstructured). In the structured networking domain using the UNSW-NB15 dataset, DP-SGD performance decreased ( Fig.\ref{fig:accuracy_comparison} and Fig.\ref{fig:perplexity_comparison}) as noise increased, whereas SA-ADP maintained good performance while still strongly preserving privacy. This flexibility highlights the scalability of SA-ADP across heterogeneous real-world scenarios 
These findings directly address the limitations outlined in the problem statement. The uniform noise of the conventional method either overprotects low-sensitive tokens at the cost of utility or under-protects high-sensitive tokens, creating privacy risks. SA-ADP overcomes this by tailoring noise to token sensitivity. 

Overall, the results suggest that moving from global to token-sensitive privacy mechanisms is a decisive step toward deployable, regulation-aligned LLM fine-tuning, particularly in PII-dense domains.

\begin{figure}[!htbp]
    \centering
    \includegraphics[width=0.90\linewidth]{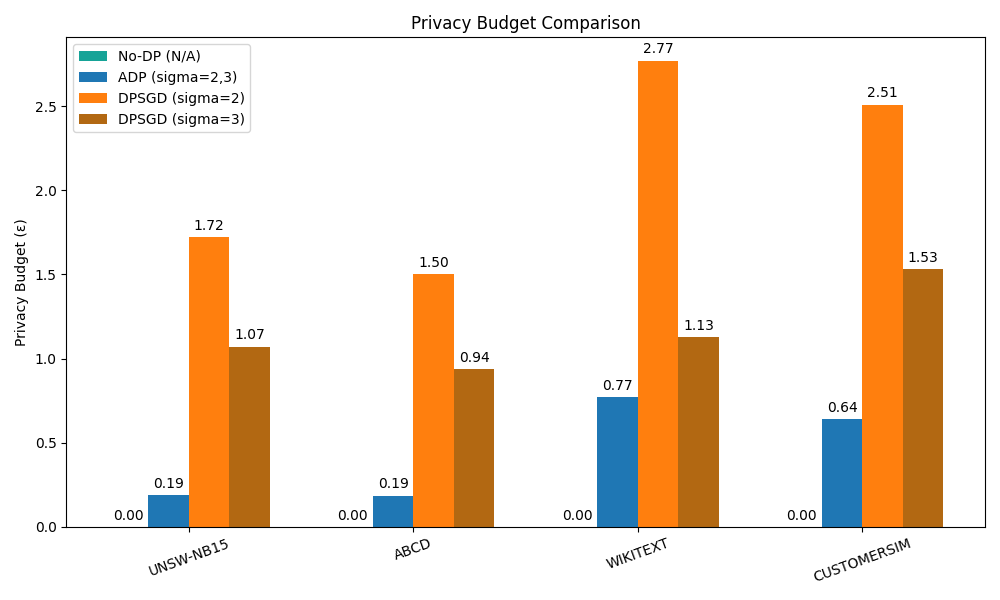}
    \caption{Privacy budget between ADP and DP-SGD across all Datasets.}
    \label{fig:privacy_budget_comparison}
\end{figure}

\section{Conclusion}
This paper introduced \textbf{Sensitivity-Aware Adaptive Differential Privacy (SA-ADP)}, a token-level mechanism that adaptively scales noise according to the sensitivity of each token. Unlike conventional DP-SGD, which applies uniform perturbation, SA-ADP aligns privacy noise with token importance, thereby achieving a tighter privacy-utility balance. Across four datasets: UNSW-NB15, ABCD, Wikitext-2, and CustomerSim, SA-ADP reduced the average privacy budget $\epsilon$ by approximately $75\%$ while maintaining about $99\%$ of baseline utility. 

These results demonstrate that a sensitivity-driven privacy control can overcome the inefficiencies of uniform noising and enable practical privacy preservation in PII-dense environments. Ultimately, SA-ADP provides a modular and adaptive framework for privacy-preserving learning that is well-suited to a range of data sensitivity regimes and regulatory requirements.

\bibliography{Reference}   
\end{document}